\documentclass{ieeeaccess}
\usepackage{amsmath,amssymb,amsfonts}
\usepackage{algorithmic}
\usepackage{graphicx}
\usepackage{textcomp}

\usepackage{algorithm}
\usepackage{subcaption}
\usepackage{multirow}
\usepackage{hhline}
\usepackage[numbers,sort&compress]{natbib}
\usepackage{soul}

\usepackage{bm}
\makeatletter
\AtBeginDocument{\DeclareMathVersion{bold}
\SetSymbolFont{operators}{bold}{T1}{times}{b}{n}
\SetSymbolFont{NewLetters}{bold}{T1}{times}{b}{it}
\SetMathAlphabet{\mathrm}{bold}{T1}{times}{b}{n}
\SetMathAlphabet{\mathit}{bold}{T1}{times}{b}{it}
\SetMathAlphabet{\mathbf}{bold}{T1}{times}{b}{n}
\SetMathAlphabet{\mathtt}{bold}{OT1}{pcr}{b}{n}
\SetSymbolFont{symbols}{bold}{OMS}{cmsy}{b}{n}
\renewcommand\boldmath{\@nomath\boldmath\mathversion{bold}}}
\makeatother

\def\BibTeX{{\rm B\kern-.05em{\sc i\kern-.025em b}\kern-.08em
    T\kern-.1667em\lower.7ex\hbox{E}\kern-.125emX}}

\begin{document}
\history{Date of publication xxxx 00, 0000, date of current version xxxx 00, 0000.}
\doi{10.1109/ACCESS.2024.0429000}

\title{Data Distribution-based Curriculum Learning}
\author{
\uppercase{Shonal Chaudhry}\authorrefmark{1}, 
\uppercase{Anuraganand Sharma}\authorrefmark{1}, \IEEEmembership{Senior Member, IEEE}}

\address[1]{School of Information Technology, Engineering, Mathematics and Physics, The University of the South Pacific, Laucala Campus, Suva, Fiji}

\markboth
{Chaudhry \headeretal: Data Distribution-based Curriculum Learning}
{Chaudhry \headeretal: Data Distribution-based Curriculum Learning}

\corresp{Corresponding author: Shonal Chaudhry (Email: shonal\_c@outlook.com).}

\begin{abstract}
The order of training samples can have a significant impact on a model's performance. Curriculum learning is an approach for gradually training a model by ordering samples from `easy' to `hard'. This paper proposes the novel idea of a curriculum learning strategy called Data Distribution-based Curriculum Learning (DDCL). DDCL uses the inherent data distribution of a dataset to build a curriculum based on the order of samples. Our proposed approach is innovative as it incorporates two distinct scoring methods known as DDCL-Density and DDCL-Point to determine the order of training samples. The DDCL-Density method assigns scores based on the density of samples favoring denser regions that can make initial learning easier. Conversely, DDCL-Point utilizes the Euclidean distance from the centroid of the dataset as a reference point to score samples providing an alternative perspective on sample difficulty. We evaluate the proposed DDCL approach by conducting experiments across various classifiers using a diverse set of small to medium-sized medical datasets. Results show that DDCL improves the classification accuracy, achieving increases ranging from 2\% to 10\% compared to baseline methods and other state-of-the-art techniques. Moreover, analysis of the error losses for a single training epoch reveals that DDCL not only improves accuracy but also increases the convergence rate, underlining its potential for more efficient training. The findings suggest that DDCL can specifically be of benefit to medical applications where data is often limited and indicate promising directions for future research in domains that involve limited datasets.
\end{abstract}


\titlepgskip=-21pt

\maketitle

\section{Introduction}
\label{Section:Introduction}
\PARstart{C}{lassification} tasks in supervised machine learning use various techniques to create classifiers suited to the problem. Some of the widely used techniques are neural networks, support vector machines (SVM) and decision trees \cite{Maxwell2018_ML_classification_remote_sensing}. Neural networks are biologically inspired computing systems \cite{Dongare2012_Intro_ANN} that are widely used as classifiers due to their ability to learn and improve their performance through the use of data and experience. SVMs use a threshold to classify a data sample as belonging to one class or another class \cite{Vapnik1999_SVM}. This is done by starting with the training data in a lower dimension and then moving to a higher dimension. The higher dimension data is then separated into groups that represent each class. Decision trees perform classification by using a tree-like structure made up of smaller decisions to make an overall decision for a given input. Multiple decision trees are often combined to create an ensemble classifier known as a Random Forest classifier \cite{Breiman2001_Random_Forests}.

The performance of these classifiers are dependent on the quality of data used and the robustness of training algorithms. Some of the factors that affect the quality of data are inaccurate, inconsistent, imbalanced, duplicate, missing and outlier samples in a dataset \cite{Gudivada2017_Data_Quality_Big_Data_Machine_Learning, Finch2012_Distribution_outlier_detection}. Studies on the quality of data have shown that these properties may lead to a significant degradation in prediction performance and cause instability in learning due to high bias and/or high variance \cite{Gudivada2017_Data_Quality_Big_Data_Machine_Learning, Taleb2018_Big_data_quality}.

To reduce the negative impact of these factors, the training algorithms applied to datasets are often designed to guide the learning model towards optimum performance \cite{Weiss2015_Structured_Training, Andrychowicz2016_Learning_to_learn_by_GD_through_GD, Zhang2018_Fine_Grained_Visual_Categorization_using_Meta_Learning}. Guidance for smaller datasets is particularly important since their limited size cannot provide the sample diversity present in larger datasets \cite{Kurdi2021_Impact_Dataset_Size}.

Gradient descent is the most common method of optimizing a neural network due to its fast convergence towards the minimum error \cite{Sharma2021_Parallel_GSGD}. Variants of gradient descent such as batch gradient descent, stochastic gradient descent (SGD) and mini-batch gradient descent exist for use with specific problem scenarios \cite{Sharma2018_GSGD}. These problems may favor computation speed, data size or a balance of speed and size as frequently seen in practice \cite{Ruder2016_An_overview_of_gradient_descent}. Support vector machines are optimized by selecting a value for the \textit{C} hyperparameter and the method of optimizing its kernel varies according to the kernel used. The Radial Basis Function (RBF) kernel is commonly used in a SVM and it is optimized by selecting the gamma ($\gamma$) hyperparameter. Both of these hyperparameters can be found using several search algorithms including trees of Parzen estimators, particle swarm optimization and Bayesian optimization. Recent findings have concluded that the first two algorithms provide better hyperparameter values with a lower execution time compared to Bayesian optimization which has a very high computational cost \cite{Wainer2021_How_to_tune_RBF_SVM_hyperparameters}. In a random forest classifier, the number of decision trees (estimators) and the number of features required to split a tree are hyperparameters that are optimized \cite{Probst2019_Hyperparameters_tuning_random_forest}. The number of estimators for a specific problem is carefully selected since it corresponds with the computation time required.

Furthermore, the data distribution plays a major role in obtaining quality predictions from classifiers. When using the same data, a particular data distribution of training data can provide better results compared to another data distribution. Research has shown that the choice of distribution can cause bias in the training data due to under-representation of the minority class \cite{Jeni2013_Imbalanced_Data_Recommendations_Performance}. This bias can have a significant impact on a model's accuracy and precision which can be critical for medical applications where training data can be limited \cite{Gyori2021_Training_data_distribution_significantly_impacts_estimation, Ma2022_Curriculum_learning_medical_report, Li2023_Curriculum_imbalanced_medical_image}. Once the distribution of data is known, a decision can be made on how data samples are selected from the dataset for optimal results. This decision may include using a specific criteria for selecting samples in a particular order. The process of selectively choosing samples from a dataset as well as determining their order is known as creating a curriculum.

Using a curriculum for improving performance of a machine learning model is known as \textit{curriculum learning} \cite{Bengio2009_Curriculum_Learning} and it is typically applied on the order of training data samples. Curriculum learning enhances training of a model by starting with simple concepts and then gradually introducing difficult concepts as training progresses \cite{Bengio2009_Curriculum_Learning, Faber2024_MNIST_ImageNet_curriculum_learning}. It is inspired by how humans are taught through curriculums in an education system by receiving basic education during childhood and then moving on to advanced education in adulthood \cite{Erstad2018_21st_Century_Curriculum, Krueger2009_Flexible_shaping_learning_small_steps}.

Curriculum learning has been applied to various problems achieving excellent results in the areas of image classification \cite{Guo2018_CurriculumNet, Zhou2021_Curriculum_Learning_Optimizing_Learning_Dynamics}, face recognition \cite{Huang2020_CurricularFace_Adaptive_CL_Loss}, visual attribute classification \cite{Sarafianos2018_Curriculum_learning_of_visual_attribute_clusters_for_multi_task_classification} and imbalanced data classification \cite{Wang2019_Dynamic_Curriculum_Learning}. The effectiveness of the `easy' to `hard' strategy, where simpler examples are presented before more complex ones, has been well-documented in literature \cite{Narvekar2020_CL_Reinforcement_Learning_Framework_Survey, Soviany2022_CL_Survey}.

However, rival approaches offer different strategies by focusing on sample difficulty and order during training. Self-paced learning (SPL) \cite{Kumar2010_Self_paced_learning}, a training strategy that also suggested presenting training samples ordered from simple to complex, built upon the concepts introduced in curriculum learning by altering the process of defining the difficulty of a sample. In SPL, the current learning progress of a model was considered for selecting the next sample whereas curriculum learning relied on a pre-determined fixed curriculum.

The concepts of SPL were used years later in a framework called self-paced curriculum learning (SPCL) \cite{Jiang2015_Self_paced_curriculum_learning}. SPCL sought to unify the ideas of curriculum learning and SPL. The authors of the study reasoned that both types of learning have drawbacks; The curriculum in curriculum learning is heavily reliant on the quality of fixed pre-determined prior knowledge and ignores feedback about the learner. In SPL, the curriculum is dynamically determined to adjust to the learning pace of the learner but is prone to over-fitting due to being unable to handle prior knowledge. According to the authors, SPCL addressed these drawbacks by introducing a flexible way of including prior knowledge while also dynamically adjusting the curriculum based on feedback from the learner.

Despite this, curriculum learning has been successfully used to improve performance of existing systems that use machine learning including unsupervised domain adaptation \cite{Choi2019_Pseudo_labeling_curriculum}, transfer learning \cite{Weinshall2018_Curriculum_learning_by_transfer_learning, Hacohen2019_On_The_Power_of_Curriculum_Learning} and reinforcement learning \cite{Narvekar2020_CL_Reinforcement_Learning_Framework_Survey, Tidd2020_Guided_CL_walking_over_complex_terrain}.

In this paper, we propose a data distribution based curriculum learning approach for classification tasks on small to medium-sized medical datasets. The proposed approach first determines the distribution of data and then information from the distribution is used to build a curriculum based on the order of samples. The approach is then evaluated on multiple datasets with classifiers based on three different types of widely used learning methods: neural networks, SVM and random forest classifier. The remainder of the paper is organized as follows. Section \ref{Section:Background} provides a background on curriculum learning as well as applications of curriculum learning to various problems. The proposed curriculum learning approach is discussed in Section \ref{Section:Proposed_Approach} and Section \ref{Section:Experiments} outlines its experimental evaluation. Section \ref{Section:Discussion} examines the experiment results whereas Section \ref{Section:Conclusion} provides concluding remarks.

\section{Background}
\label{Section:Background}
The concept of using a curriculum in machine learning was introduced by Bengio et al. in 2009 \cite{Bengio2009_Curriculum_Learning}. Their work defined curriculum learning as a training strategy where fewer simple concepts are presented in order at the beginning with greater complex concepts introduced at later stages. The authors conducted experiments on shape recognition and language modelling for scenarios with and without a curriculum. Experiment results showed that by using a curriculum, faster training and better convergence can be achieved.

Hacohen et al. \cite{Hacohen2019_On_The_Power_of_Curriculum_Learning} expanded the definition of curriculum learning by describing it as consisting of two tasks with specific functions, termed the \textit{scoring function} and the \textit{pacing function}. The scoring function ranks the difficulty of the samples in a curriculum whereas the pacing function determines how often newer samples are presented to a model during training.

Curriculum learning has been applied to a variety of problems achieving promising results. Some of these problems include image classification \cite{Guo2018_CurriculumNet, Zhou2021_Curriculum_Learning_Optimizing_Learning_Dynamics}, face recognition \cite{Huang2020_CurricularFace_Adaptive_CL_Loss}, visual attribute classification \cite{Sarafianos2018_Curriculum_learning_of_visual_attribute_clusters_for_multi_task_classification} and imbalanced data classification \cite{Wang2019_Dynamic_Curriculum_Learning}. Moreover, curriculum learning has also been successfully used to improve performance of systems where a curriculum is typically not considered. These include unsupervised domain adaptation \cite{Choi2019_Pseudo_labeling_curriculum}, transfer learning \cite{Weinshall2018_Curriculum_learning_by_transfer_learning, Hacohen2019_On_The_Power_of_Curriculum_Learning}, bipedal walking for robots \cite{Tidd2020_Guided_CL_walking_over_complex_terrain} and medical report generation \cite{Ma2022_Curriculum_learning_medical_report}.

Overall, we categorize the reviewed curriculum learning literature into three different groups: the applications of standard curriculum learning \cite{Sarafianos2018_Curriculum_learning_of_visual_attribute_clusters_for_multi_task_classification, Tidd2020_Guided_CL_walking_over_complex_terrain, Ma2022_Curriculum_learning_medical_report}, variations of learning styles \cite{Weinshall2018_Curriculum_learning_by_transfer_learning, Hacohen2019_On_The_Power_of_Curriculum_Learning, Huang2020_CurricularFace_Adaptive_CL_Loss, Zhou2021_Curriculum_Learning_Optimizing_Learning_Dynamics} and research that makes use of the data density \cite{Guo2018_CurriculumNet, Choi2019_Pseudo_labeling_curriculum, Wang2019_Dynamic_Curriculum_Learning}.

\subsection{Standard curriculum learning}
Sarafianos et al. \cite{Sarafianos2018_Curriculum_learning_of_visual_attribute_clusters_for_multi_task_classification} applied curriculum learning to visual attribute classification by introducing a method that combined curriculum learning with multi-task learning. Their method performed end-to-end learning by providing a convolutional neural network (CNN) with a complete image of a human without additional data to aid in classification. Then in multi-task learning, the tasks were split into groups using clustering. Curriculum learning was applied to the groups starting with the highest within-group cross-correlation and moving to the less correlated ones. Experiments performed on three datasets consisting of humans standing resulted in an increase in performance by up to 10\%.

Tidd et al. \cite{Tidd2020_Guided_CL_walking_over_complex_terrain} use curriculum learning to train deep reinforcement learning policies for bipedal walking of a robot over challenging terrain. The authors created an easy to hard curriculum using a three stage framework: In the first stage, guiding forces were applied to the joints and the base of the robot to start learning on easy terrain and gradually increased to difficult terrain. At the second stage when the terrain was most difficult, the guiding forces applied to the robot were slowly decreased. During the final stage, the magnitude of external random perturbations were increased to improve the robustness of the policy. Simulation experiments conducted by the authors demonstrated that a curriculum approach was effective in learning to walk for five types of terrain.
\begin{figure*}[htb]
    \centering
    \includegraphics[scale = 0.62]{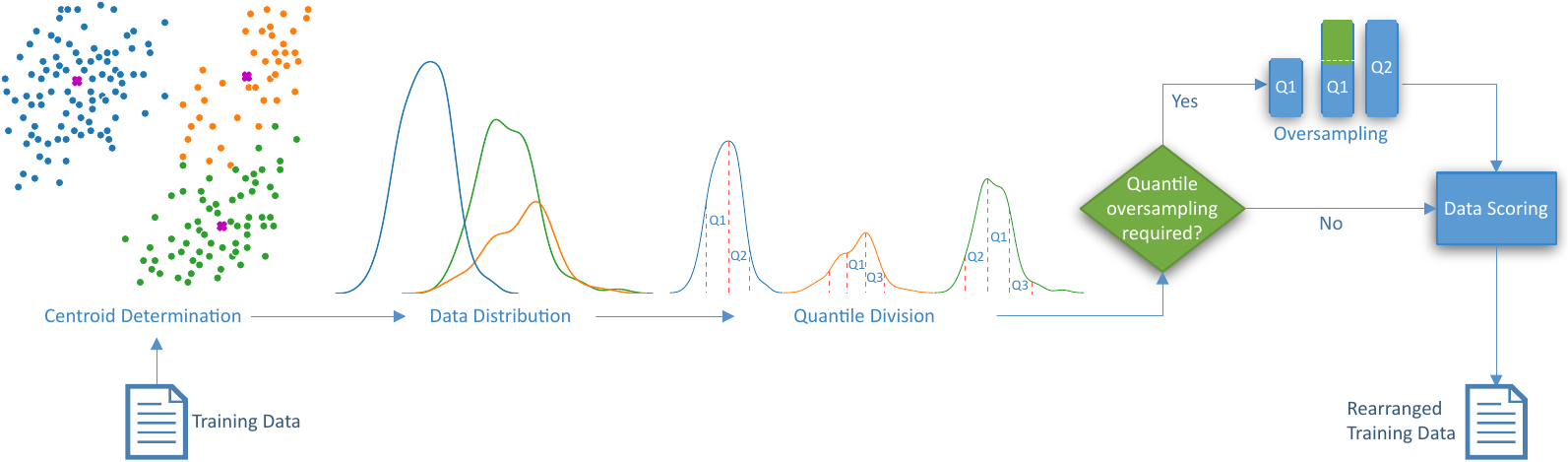}
    \caption{Data Distribution-based Curriculum Learning}
    \label{Fig:Data_Distribution_CL}
\end{figure*}

Ma et al. \cite{Ma2022_Curriculum_learning_medical_report} applied curriculum learning to the medical report generation task. Their work introduced a framework capable of learning medical reports from limited medical data while reducing data bias. The framework learned in an easy to hard manner using a two step process where at first simple reports were used and then gradually complex reports consisting of rare and diverse medical abnormalities were attempted. This process effectively simulated the learning process of radiologists. The method was evaluated on two public datasets resulting in a boost in performance of the baselines.

\subsection{Learning style variations}
In \cite{Weinshall2018_Curriculum_learning_by_transfer_learning}, the Weinshall et al. used transfer learning to implement curriculum learning. They presented an approach where the curriculum was inferred through transfer learning from another network that was pre-trained on another task. A CNN with two architectures trained using curriculum learning was used to evaluate the proposed method on the CIFAR-100 \cite{Krizhevsky2009_CIFAR} and STL-10 datasets \cite{Coates2011_STL_10}. Experiment results from their work concluded that with curriculum learning the convergence is faster during the beginning of training and improved generalization is achieved when more difficult tasks are used.

Hacohen et al. applied curriculum learning to the training of deep networks \cite{Hacohen2019_On_The_Power_of_Curriculum_Learning}. They used transfer learning and bootstrapping as two different techniques for the easy to hard process which they termed the \textit{scoring function}. With transfer learning, a `teacher' network was trained on a large dataset and its prediction performance was used as the scoring function. In bootstrapping, a network was trained without a curriculum and its performance on the training data defined the scoring function. Then, the network was retrained from scratch using curriculum learning. For the \textit{pacing function}, three approaches were used with all pacing functions having comparable performance based on experiments done by the authors. Experiments done on six test cases showed that using a curriculum provided high accuracy and better convergence with the amount of improvement ranging from small to large depending on the test case.

Huang et al. \cite{Huang2020_CurricularFace_Adaptive_CL_Loss} propose a different approach to curriculum learning in the area of face recognition. Instead of using a traditional curriculum created by fixed ordering of samples with increasing difficulty, the authors introduce a loss function which incorporates adaptive curriculum learning. During the training process, samples are randomly selected for each mini-batch and the curriculum is created adaptively from the selected samples. Furthermore, the definition of hard samples is dynamic with a sample classified as hard during the start of training becoming easy towards the end. Overall, the loss function emphasizes easy samples at the start and hard samples later. The authors conducted experiments on benchmark data achieving improved performance over other methods.

Recent research has looked at improving the general curriculum learning process. Zhou et al. \cite{Zhou2021_Curriculum_Learning_Optimizing_Learning_Dynamics} presented a curriculum learning method known as dynamics-optimized curriculum learning (DoCL). DoCL selected training samples at each step using weighted sampling based on the scores. The authors conducted experiments on more than nine datasets achieving results that significantly improved the performance and efficiency compared to existing curriculum learning methods.

\subsection{Data density based curriculum learning} 
Our review of literature for curriculum learning has found few papers that utilize the density of the data. These works are presented in this section and our work on DDCL adds to this field of research. Specifically, DDCL addresses the key limitation of other works \cite{Guo2018_CurriculumNet, Choi2019_Pseudo_labeling_curriculum} that require large datasets to utilize data density by effectively operating on small datasets. Moreover, clustering in DDCL is used to determine the class centroids rather than the density compared to other works.

Researchers in \cite{Guo2018_CurriculumNet} presented a method for training deep neural networks on large-scale weakly-supervised web images. Their training strategy utilized curriculum learning to effectively handle the large amount of noisy labels and data imbalance during the training process. They proposed a curriculum created by measuring the complexity of data using the density of its data. Experiments using the proposed training strategy resulted in state-of-the-art performance on benchmark datasets.

Choi et al. \cite{Choi2019_Pseudo_labeling_curriculum} presented another method of using the density of data to generate the curriculum. Their method applied clustering on the data where higher density samples were considered simple and lower density samples were considered as complex. The proposed method was robust against false pseudo-labelled samples due to the use of a pseudo-labelling curriculum. They achieved state-of-the-art classification results on three benchmark datasets.

Curriculum learning has been applied to imbalanced data classification as well. Dynamic Curriculum Learning (DCL) proposed by Wang et.al \cite{Wang2019_Dynamic_Curriculum_Learning} addressed the problem of requiring prior knowledge to train a system when using conventional techniques. DCL used a two-level curriculum scheduler made up of a sampling scheduler and a loss scheduler. The sampling scheduler found the best samples in a batch to train the model by dynamically managing the target data distributions from imbalanced to balanced and from easy to hard. The loss scheduler controlled the learning weights between classification loss and metric learning loss. Experiments achieved state-of-the-art performance on the CelebA face attribute dataset and the RAP pedestrian attribute dataset.
\begin{figure*}[ht]
    \begin{subfigure}[b]{0.5\textwidth}
        \includegraphics[scale = 0.75]{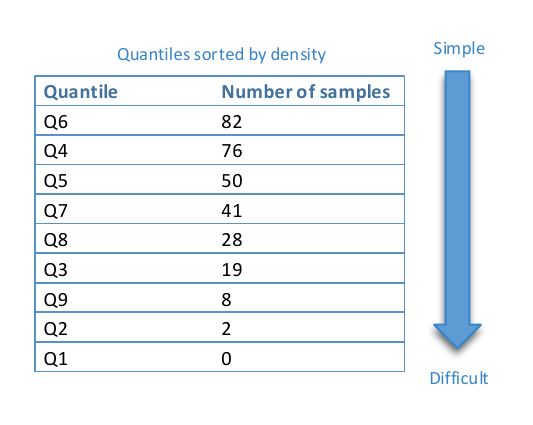}
        \caption{Density scoring showing the densities of each quantile.}
        \label{Fig:DDCL_Density}
    \end{subfigure}
    \begin{subfigure}[b]{0.475\textwidth}
        \includegraphics[scale = 0.71]{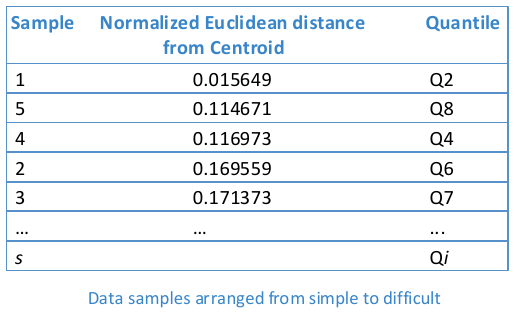}
        \caption{Point scoring showing samples sorted by the $\hat{E_s}$ value.}
        \label{Fig:DDCL_Point}
    \end{subfigure}
    \caption{Example of DDCL scoring methods on Haberman's Survival data.}
    \label{Fig:Scoring}
\end{figure*}

\begin{algorithm}
    \caption{DDCL algorithm}
    \label{Alg:DDCL}
        \begin{algorithmic}[1]
            \STATE \textbf{Variables:}
            \STATE Classes: $S$
            \STATE Data size: $N$
            \STATE $N_s$ - dataset for class $s$
            \STATE $|N_s|$ - data size (cardinality)
            \STATE $h$ - bandwidth
            \FORALL{$s \in S$}
                \STATE ${O_s} \gets \sum_{n=0}^{|N_s|} \min_{\mu_s\in S} (\parallel x_{n}-\mu_{s}\parallel^2) $
                \STATE $s \gets s \cup O_s$
                \STATE $E_s \gets \varnothing$
                \FORALL{$x \in s$}
                    \STATE $E_s \gets E_s \cup \parallel{O_{s} - x}\parallel$
                \ENDFOR
                \STATE $D_s \gets \frac{1}{|N_s|h} \sum_{i=1}^{|N_s|} K \left(\frac{y - \hat{E}_i}{h}\right)$ \COMMENT{\begin{small}Kernel Density Estimation\end{small}}
                \FOR{$q \in Q^s$}
                    \IF{$|q| < 3$}
                        \STATE $q \gets SMOTE(q)$
                    \ENDIF
                \ENDFOR
                \IF{Scoring type $\gets$ ``Density''}
                    \STATE go to Algorithm \ref{Alg:Density}
                \ELSIF{Scoring type $\gets$ ``Point''}
                    \STATE go to Algorithm \ref{Alg:Point}
                \ENDIF
                \STATE \RETURN{$T \gets R_s$}
            \ENDFOR
        \end{algorithmic}
\end{algorithm}

\section{Proposed Curriculum Learning Approach}
\label{Section:Proposed_Approach}
The curriculum learning approach proposed in this paper is based on the data distribution of a dataset. This is referred to as \textit{Data Distribution-based Curriculum Learning (DDCL)} and involves multiple steps as visualized through Figure \ref{Fig:Data_Distribution_CL}. It begins with dividing the data into groups based on their target classes. Then, the centroids for each group are calculated (Centroid Determination) and used to compute the Euclidean distance between each sample and its centroid. The data distribution for each class is determined next and utilized to divide the data for individual classes into quantiles (Q1, Q2, Q3). After quantile division, oversampling is optionally performed on quantiles where the number of samples may be highly imbalanced compared to other quantiles. Lastly, the data is scored with either density or point scoring and rearranged for use in training.

The DDCL process is further detailed in Algorithm \ref{Alg:DDCL}, Algorithm \ref{Alg:Density} and Algorithm \ref{Alg:Point}. In the first step of DDCL, the training data samples are grouped according to their classes ($S$) with the number of groups being equal to the number of unique classes in the data. Next, the grouped training data is used to calculate the centroids ($O_s$) for each class by performing clustering on each data group. Then, the centroids are used to compute the Euclidean distance from the relevant centroid ($E_s$) for each data sample ($x$). The third step determines the data distribution for a class ($D_s$) by plotting the normalized values of the Euclidean distances. Once the distribution of the training data is known, the data samples are divided into quantiles ($Q^s$). Each quantile ($q$) is then examined for the number of samples and the quantiles with the lowest number of samples are oversampled using SMOTE \cite{Chawla2002_SMOTE} if there are sufficient samples. SMOTE creates synthetic samples using the original data and different variants of it have been developed \cite{Sharma2022_SMOTified_GAN} since the initial method was proposed. The initial method of SMOTE is used in our proposed approach to address the potential lack of samples in a given quantile. Otherwise, no oversampling is applied due to insufficient samples. After the samples are divided into quantiles and optionally oversampled, each data sample is scored using either density or point scoring and rearranged accordingly ($R_s$). Finally, the rearranged training data ($T$) is ready to be passed to one of the learning methods for training and evaluation.

DDCL consists of two types of scoring methods: a sample density based method and a Euclidean distance based method. These are referred to as DDCL-Density and DDCL-Point respectively and are detailed in the following sections.
\begin{table}[ht]
  \caption{Datasets used for evaluating DDCL.}
  \label{Table:Datasets}
  \centering
    \resizebox{8.5cm}{!}
    {
    \begin{tabular}{lccc}
      \hline
      \textbf{Dataset} & \multirow{2}{1.2cm}{\textbf{Data Instances}} & \textbf{Attributes} & \textbf{Class Type} \\
      & & & \\
      \hline
      \multirow{2}{2.7cm}{Breast Cancer Wisconsin (Diagnostic)} & \multirow{2}{*}{569} & \multirow{2}{*}{31} & \multirow{2}{*}{Binary} \\
      & & & \\
      Cancer & 457 & 9 & Binary \\
      Haberman's Survival & 306 & 3 & Binary \\
      Liver Disorder & 345 & 6 & Binary \\
      Pima Indians Diabetes & 768 & 8 & Binary \\
      New-Thyroid & 215 & 5 & Multi-class \\
      Diabetes 130 \cite{Shane2018_Readmission_Prediction} & 86556 & 46 & Multi-class \\
      \hline
    \end{tabular}
    }
\end{table}

\subsection{Density based DDCL}
Figure \ref{Fig:DDCL_Density} illustrates the DDCL-Density process where the scoring is applied to the quantiles based on the number of samples in the quantile. Quantiles with the greatest number of samples (Q6) are given a higher score whereas quantiles with the least number of samples (Q1) are given a lower score. Higher scores are considered simple and lower scores are considered difficult. This results in a final quantile order sorted from the highest to lowest density thus defining the curriculum to be used during training.
\begin{table}[ht]
  \caption{Neural Network results with and without DDCL.}
  \label{Table:NN_results}
  \centering
    \resizebox{8.5cm}{!}
    {
    \begin{tabular}{llccc}
      \hline
      \textbf{Dataset} & \textbf{Test Scenario} & \textbf{Worst \%} & \textbf{Best \%} & \textbf{Average $\pm \sigma$} \\
      \hhline{=====}
      \multirow{3}{1.6cm}{Breast Cancer (Diagnostic)} & No Curriculum & 91.304 & 100.000 & 96.232 $\pm$2.442 \\
      & DDCL-Density & 89.130 & 100.000 & 96.594 $\pm$2.673 \\
      & \textbf{DDCL-Point} & \textbf{86.957} & \textbf{100.000} & \textbf{97.319 $\pm$2.953} \\
      \hline
      \multirow{3}{1.6cm}{Cancer} & No Curriculum & 86.486 & 100.000 & 94.955 $\pm$3.993 \\
      & \textbf{DDCL-Density} & \textbf{91.667} & \textbf{100.000} & \textbf{96.852 $\pm$2.658} \\
      & DDCL-Point & 86.111 & 100.000 & 96.574 $\pm$3.100 \\
      \hline
      \multirow{3}{1.6cm}{Haberman's Survival} & No Curriculum & 45.833 & 79.167 & 65.833 $\pm$8.898 \\
      & \textbf{DDCL-Density} & \textbf{50.000} & \textbf{87.500} & \textbf{69.167 $\pm$10.353} \\
      & DDCL-Point & 41.667 & 87.500 & 67.222 $\pm$9.424 \\
      \hline
      \multirow{3}{1.6cm}{Liver Disorder} & No Curriculum & 57.143 & 85.714 & 69.048 $\pm$6.477 \\
      & DDCL-Density & 46.429 & 85.714 & 68.214 $\pm$9.052 \\
      & \textbf{DDCL-Point} & \textbf{57.143} & \textbf{82.143} & \textbf{69.524 $\pm$6.098} \\
      \hline
      \multirow{3}{1.6cm}{Pima Indians Diabetes} & No Curriculum & 54.098 & 86.885 & 71.803 $\pm$6.658 \\
      & \textbf{DDCL-Density} & \textbf{55.738} & \textbf{85.246} & \textbf{72.732 $\pm$5.933} \\
      & DDCL-Point & 57.377 & 80.328 & 71.803 $\pm$5.460 \\
      \hline
      \multirow{3}{1.6cm}{New-Thyroid} & No Curriculum & 82.353 & 100.000 & 93.137 $\pm$5.280 \\
      & \textbf{DDCL-Density} & \textbf{82.353} & \textbf{100.000} & \textbf{95.490 $\pm$5.409} \\
      & DDCL-Point & 82.353 & 100.000 & 94.118 $\pm$5.683 \\
      \hline
      \multirow{3}{1.6cm}{Diabetes 130} & No Curriculum & 53.235 & 54.694 & 54.029 $\pm$0.518 \\
      & \textbf{DDCL-Density} & \textbf{56.148} & \textbf{56.798} & \textbf{56.579 $\pm$0.246} \\
      & DDCL-Point & 54.400 & 56.278 & 55.139 $\pm$0.696 \\
      \hline
    \end{tabular}
    }
\end{table}
\begin{table}[ht]
  \caption{SVM results with and without DDCL.}
  \label{Table:SVM_results}
  \centering
    \resizebox{8.5cm}{!}
    {
    \begin{tabular}{llccc}
      \hline
      \textbf{Dataset} & \textbf{Test Scenario} & \textbf{Worst \%} & \textbf{Best \%} & \textbf{Average $\pm \sigma$} \\
      \hhline{=====}
      \multirow{3}{1.6cm}{Breast Cancer (Diagnostic)} & No Curriculum & 94.737 & 99.415 & 97.700 $\pm$1.188 \\
      & \textbf{DDCL-Density} & \textbf{95.322} & \textbf{99.415} & \textbf{97.758 $\pm$1.025} \\
      & DDCL-Point & 94.737 & 99.415 & 97.719 $\pm$1.239 \\
      \hline
      \multirow{3}{1.6cm}{Cancer} & No Curriculum & 94.928 & 99.275 & 96.715 $\pm$1.134 \\
      & \textbf{DDCL-Density} & \textbf{94.118} & \textbf{99.265} & \textbf{96.985 $\pm$1.145} \\
      & DDCL-Point & 92.647 & 99.265 & 96.225 $\pm$1.562 \\
      \hline
      \multirow{3}{1.6cm}{Haberman's Survival} & No Curriculum & 65.217 & 78.261 & 72.609 $\pm$4.051 \\
      & \textbf{DDCL-Density} & \textbf{65.217} & \textbf{83.696} & \textbf{73.841 $\pm$4.836} \\
      & DDCL-Point & 67.391 & 79.348 & 73.080 $\pm$3.329 \\
      \hline
      \multirow{3}{1.6cm}{Liver Disorder} & No Curriculum & 44.231 & 76.923 & 63.750 $\pm$7.264 \\
      & DDCL-Density & 49.038 & 75.000 & 66.699 $\pm$5.515 \\
      & \textbf{DDCL-Point} & \textbf{55.769} & \textbf{75.962} & \textbf{68.558 $\pm$4.783} \\
      \hline
      \multirow{3}{1.6cm}{Pima Indians Diabetes} & No Curriculum & 71.429 & 80.519 & 76.046 $\pm$2.500 \\
      & \textbf{DDCL-Density} & \textbf{70.996} & \textbf{82.251} & \textbf{77.128 $\pm$2.991} \\
      & DDCL-Point & 72.294 & 82.684 & 76.508 $\pm$2.514 \\
      \hline
      \multirow{3}{1.6cm}{New-Thyroid} & No Curriculum & 92.308 & 100.000 & 97.128 $\pm$2.472 \\
      & \textbf{DDCL-Density} & \textbf{92.308} & \textbf{100.000} & \textbf{97.282 $\pm$1.892} \\
      & DDCL-Point & 92.308 & 100.000 & 96.872 $\pm$1.563 \\
      \hline
      \multirow{3}{1.6cm}{Diabetes 130} & No Curriculum & 64.285 & 65.487 & 64.948 $\pm$0.302 \\
      & DDCL-Density & 64.220 & 65.457 & 64.977 $\pm$0.298 \\
      & \textbf{DDCL-Point} & \textbf{64.301} & \textbf{65.469} & \textbf{65.052 $\pm$0.266} \\
      \hline
    \end{tabular}
    }
\end{table}
\begin{table}[ht]
  \caption{Random Forest results with and without DDCL.}
  \label{Table:RF_results}
  \centering
    \resizebox{8.5cm}{!}
    {
    \begin{tabular}{llccc}
      \hline
      \textbf{Dataset} & \textbf{Test Scenario} & \textbf{Worst \%} & \textbf{Best \%} & \textbf{Average $\pm \sigma$} \\
      \hhline{=====}
      \multirow{3}{1.6cm}{Breast Cancer (Diagnostic)} & No Curriculum & 94.152 & 98.246 & 95.945 $\pm$1.034 \\
      & DDCL-Density & 92.398 & 98.246 & 95.653 $\pm$1.581 \\
      & \textbf{DDCL-Point} & \textbf{91.228} & \textbf{98.246} & \textbf{95.945 $\pm$1.517} \\
      \hline
      \multirow{3}{1.6cm}{Cancer} & No Curriculum & 92.754 & 100.000 & 97.150 $\pm$1.374 \\
      & DDCL-Density & 94.853 & 100.000 & 96.691 $\pm$1.298 \\
      & \textbf{DDCL-Point} & \textbf{94.118} & \textbf{100.000} & \textbf{97.206 $\pm$1.361} \\
      \hline
      \multirow{3}{1.6cm}{Haberman's Survival} & No Curriculum & 58.696 & 76.087 & 68.261 $\pm$4.100 \\
      & \textbf{DDCL-Density} & \textbf{65.217} & \textbf{79.348} & \textbf{70.145 $\pm$3.825} \\
      & DDCL-Point & 57.609 & 77.174 & 69.746 $\pm$4.367 \\
      \hline
      \multirow{3}{1.6cm}{Liver Disorder} & No Curriculum & 59.615 & 78.846 & 70.417 $\pm$5.104 \\
      & \textbf{DDCL-Density} & \textbf{67.308} & \textbf{77.885} & \textbf{72.917 $\pm$2.709} \\
      & DDCL-Point & 60.577 & 78.846 & 70.064 $\pm$4.077 \\
      \hline
      \multirow{3}{1.6cm}{Pima Indians Diabetes} & No Curriculum & 71.861 & 79.654 & 75.483 $\pm$1.985 \\
      & DDCL-Density & 70.996 & 80.087 & 75.339 $\pm$2.200 \\
      & \textbf{DDCL-Point} & \textbf{65.368} & \textbf{81.818} & \textbf{76.003 $\pm$3.461} \\
      \hline
      \multirow{3}{1.6cm}{New-Thyroid} & No Curriculum & 95.385 & 100.000 & 97.949 $\pm$1.214 \\
      & \textbf{DDCL-Density} & \textbf{92.308} & \textbf{100.000} & \textbf{98.000 $\pm$1.911} \\
      & DDCL-Point & 95.385 & 100.000 & 97.897 $\pm$1.512 \\
      \hline
      \multirow{3}{1.6cm}{Diabetes 130} & No Curriculum & 65.830 & 66.658 & 66.218 $\pm$0.228 \\
      & \textbf{DDCL-Density} & \textbf{65.626} & \textbf{66.948} & \textbf{66.242 $\pm$0.309} \\
      & DDCL-Point & 65.542 & 66.805 & 66.199 $\pm$0.293 \\
      \hline
    \end{tabular}
    }
\end{table}

\begin{algorithm}
    \caption{DDCL Density}
    \label{Alg:Density}
        \begin{algorithmic}[1]
            \STATE $C_s \gets \varnothing$ \COMMENT{Initialize}
            \FORALL{$q \in Q^s$}
                \STATE $C_s \gets C_s \cup |q|$
            \ENDFOR
            \STATE $R_s \gets sort(\forall \: C_s \mid s \in S)$
            \STATE \RETURN {$R_s$}
        \end{algorithmic}
\end{algorithm}

Algorithm \ref{Alg:Density} provides details of the DDCL-Density scoring process. It shows that the process starts with determining the density of each quantile by taking their cardinality, $|q|$. Once the densities for all quantiles ($C_s$) are known, the training data is sorted according to their quantile's density from highest to lowest. Finally, the rearranged data, $R_s$, is ready for use in training.

\subsection{Point based DDCL}
On the other hand, Figure \ref{Fig:DDCL_Point} illustrates how DDCL-Point assigns a score to the individual data samples in the dataset by utilizing their normalized Euclidean distance from centroid rather than examining quantiles. In this scoring method, data samples with the shorter Euclidean distances (1 and 5) are given the highest scores whereas samples with the longer Euclidean distances (3 and $s$) are assigned the lowest scores. As with DDCL-Density, higher scores in this point scoring method are considered simple whereas lower scores are considered difficult. Using DDCL-Point results in a curriculum where the training samples are ordered based on their individual characteristic rather than as a group.
\begin{algorithm}
    \caption{DDCL Point}
    \label{Alg:Point}
        \begin{algorithmic}[1]
            \STATE $R_s \gets sort(\forall \: \hat{E_s} | s \in S)$
            \STATE \RETURN{$R_s$}
        \end{algorithmic}
\end{algorithm}

Algorithm \ref{Alg:Point} explains the DDCL-Point scoring method and it shows that the data samples are sorted using the normalized Euclidean distance from centroid, $\hat{E_s}$. The sorting is done from the shortest distance to the longest regardless of which quantile a sample is assigned to. For example, consider a dataset with four quantiles with each quantile having at least five data samples. The first sample with the lowest $\hat{E_s}$ value may belong to quantile 1. Then, the sample with the second lowest $\hat{E_s}$ could come from quantile 4. The third lowest $\hat{E_s}$ value can be a sample from quantile 2. Next, the fourth lowest $\hat{E_s}$ may come from a sample in quantile 3. This method of sorting then repeats until $\hat{E_s}$ values from all data samples are accounted for with the resulting order of quantiles and samples ($R_s$) becoming randomized in contrast to DDCL-Density.
\begin{figure*}[ht]
    \centering
    \includegraphics[scale = 0.50]{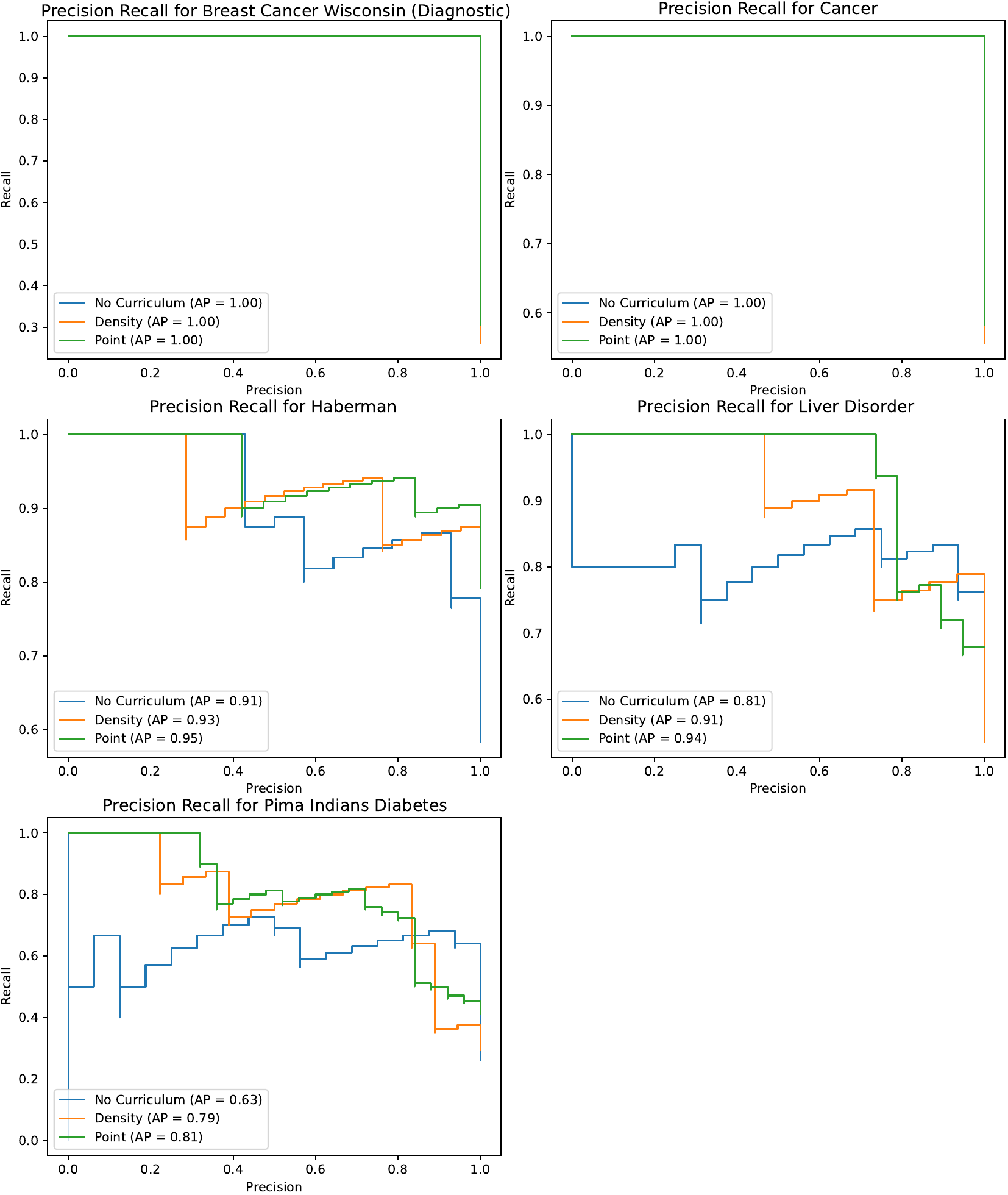}
    \caption{Precision-Recall curves for binary classification datasets.}
    \label{Fig:Precision_Recall}
\end{figure*}
\begin{figure*}
    \centering
    \includegraphics[scale = 0.50]{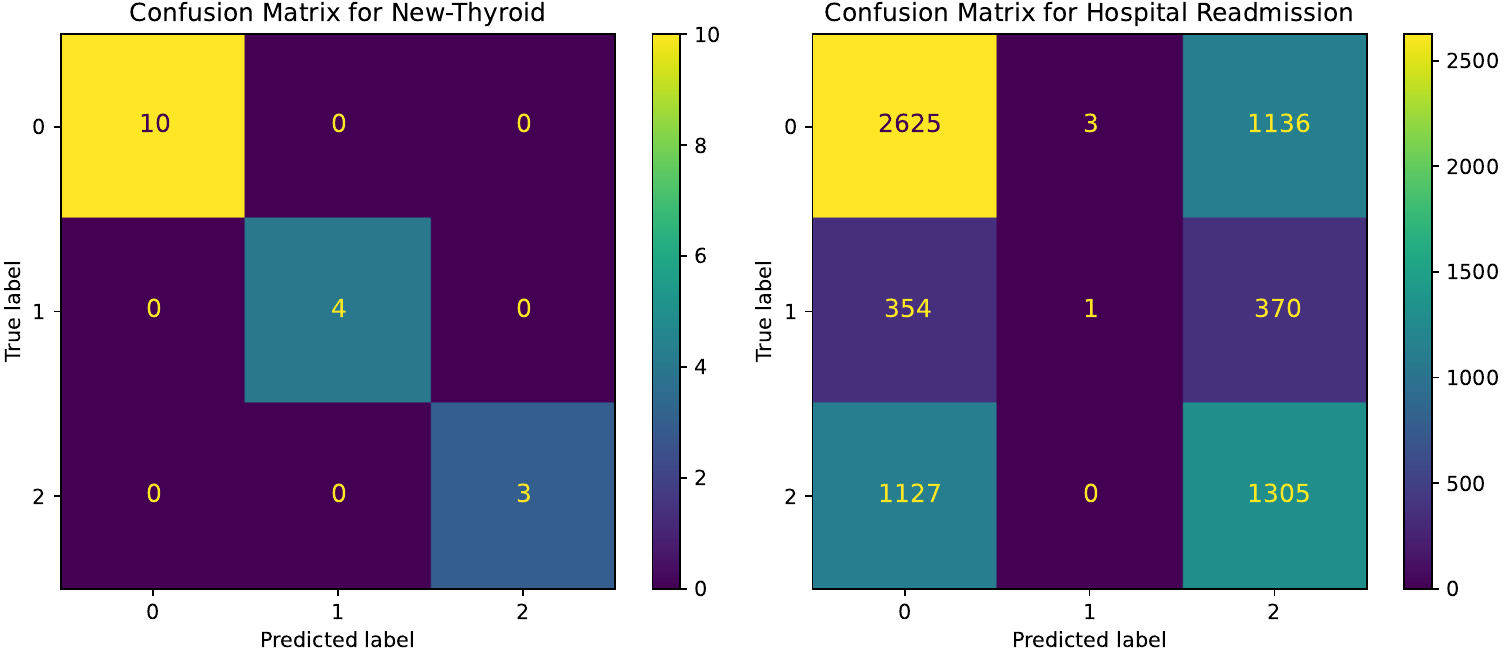}
    \caption{Confusion matrix for multi-class classification datasets.}
    \label{fig:Confusion_Matrix}
\end{figure*}

\section{Experiments}
\label{Section:Experiments}
DDCL is evaluated by performing classification tasks on seven datasets obtained from the UCI Machine Learning Repository \cite{Dua2019_UCI_ML_Repository}. As listed in Table \ref{Table:Datasets}, the datasets used for the experiments focus on small to medium-sized medical data. Three types of classifiers are used for evaluating DDCL; neural network, SVM and random forest.
\begin{table*}[ht]
  \caption{Comparison of DDCL results with other works.}
  \label{Table:DDCL_Compare}
  \centering
    \resizebox{11cm}{!}
    {
    \begin{tabular}{llcc}
      \hline
      \textbf{Dataset} & \textbf{Method} & \textbf{Year} & \textbf{Accuracy \%} \\
      \hline
      \multirow{2}{2.7cm}{Breast Cancer Wisconsin (Diagnostic)} & SVM \cite{Naji2021_Cancer_Diagnostic} & 2021 & 97.20 \\
      & \textbf{SVM (DDCL)} & \textbf{2024} & \textbf{99.42} \\
      \hline
      \multirow{2}{*}{Haberman's Survival} & Random Forest \cite{Zakaria2023_Pima_Diabetes_Haberman} & 2023 & 74.00 \\
      & \textbf{Random Forest (DDCL)} & \textbf{2024} & \textbf{79.35} \\
      \hline
      \multirow{2}{*}{Liver Disorder} & Neural Network \cite{Kumar2023_Liver_Disorder} & 2023 & 76.67 \\
      & \textbf{Neural Network (DDCL)} & \textbf{2024} & \textbf{82.14} \\
      \hline
      \multirow{2}{*}{Pima Indians Diabetes} & Random Forest \cite{Chang2022_Pima_Diabetes} & 2022 & 79.57 \\
      & Random Forest \cite{Zakaria2023_Pima_Diabetes_Haberman} & 2023 & 79.00 \\
      & \textbf{Random Forest (DDCL)} & \textbf{2024} & \textbf{81.82} \\
      \hline
      \multirow{2}{*}{Diabetes 130} & Random Forest \cite{Shane2018_Readmission_Prediction} & 2018 & 55.97 \\
      & \textbf{Random Forest (DDCL)} & \textbf{2024} & \textbf{66.95} \\
      \hline
    \end{tabular}
    }
\end{table*}

The neural network architecture used for evaluating DDCL consists of one input layer, a variable number of hidden layers and one output layer. The size of the input layer is equal to the number of attributes in the dataset and varies depending on the selected dataset. Similarly, the number of hidden layers changes based on the dataset and is a hyperparameter that can be tuned. Bayesian optimization is used on the model to tune the hidden layer values. Finally, the size of the output layer can be configured for binary or multi-class classification. The SVM classifier for evaluating DDCL uses a \textit{C} value of 1.0 with a RBF kernel. The gamma ($\gamma$) value of the RBF kernel scales according to the dataset used. A RBF kernel is chosen due to its good generalization ability and robustness to input noise. The random forest classifier in our evaluation of DDCL uses 100 estimators and a value of 2 features for splitting a node.

During the neural network experiments, the selected dataset is split into training, validation and testing subsets. 20{\%} of the data is used for validation, 10{\%} for testing and the remainder for training. The neural network model is then trained on the training subset for 200 epochs and tested on the testing subset. Likewise in the SVM and random forest experiments, the selected dataset is split into training and testing subsets with 70{\%} and 30{\%} of the data being used respectively. The SVM and random forest classifiers are trained using the training subset of the data and tested on the testing subset.

For each dataset, 30 experiments runs are performed for each of the three test scenarios: \textit{No Curriculum}, \textit{DDCL-Density} and \textit{DDCL-Point}. The same number of experiment runs is used for all three classifiers. The first \textbf{No Curriculum} test scenario takes the input data and uses it to train a neural network, SVM and random forest classifier without altering the order of the training data. After training, classification experiments are conducted on the testing subset of data. In the second \textbf{DDCL-Density} test scenario, the input data is processed using the steps outlined in the DDCL method and scored by density. Classification is then performed on the testing subset as stated earlier. The \textbf{DDCL-Point} test scenario processes the input data through DDCL with the point scoring approach being used for the scoring step. Experiments for classification are likewise performed on the testing subset.
\begin{figure*}[ht]
    \centering
    \includegraphics[scale = 0.50]{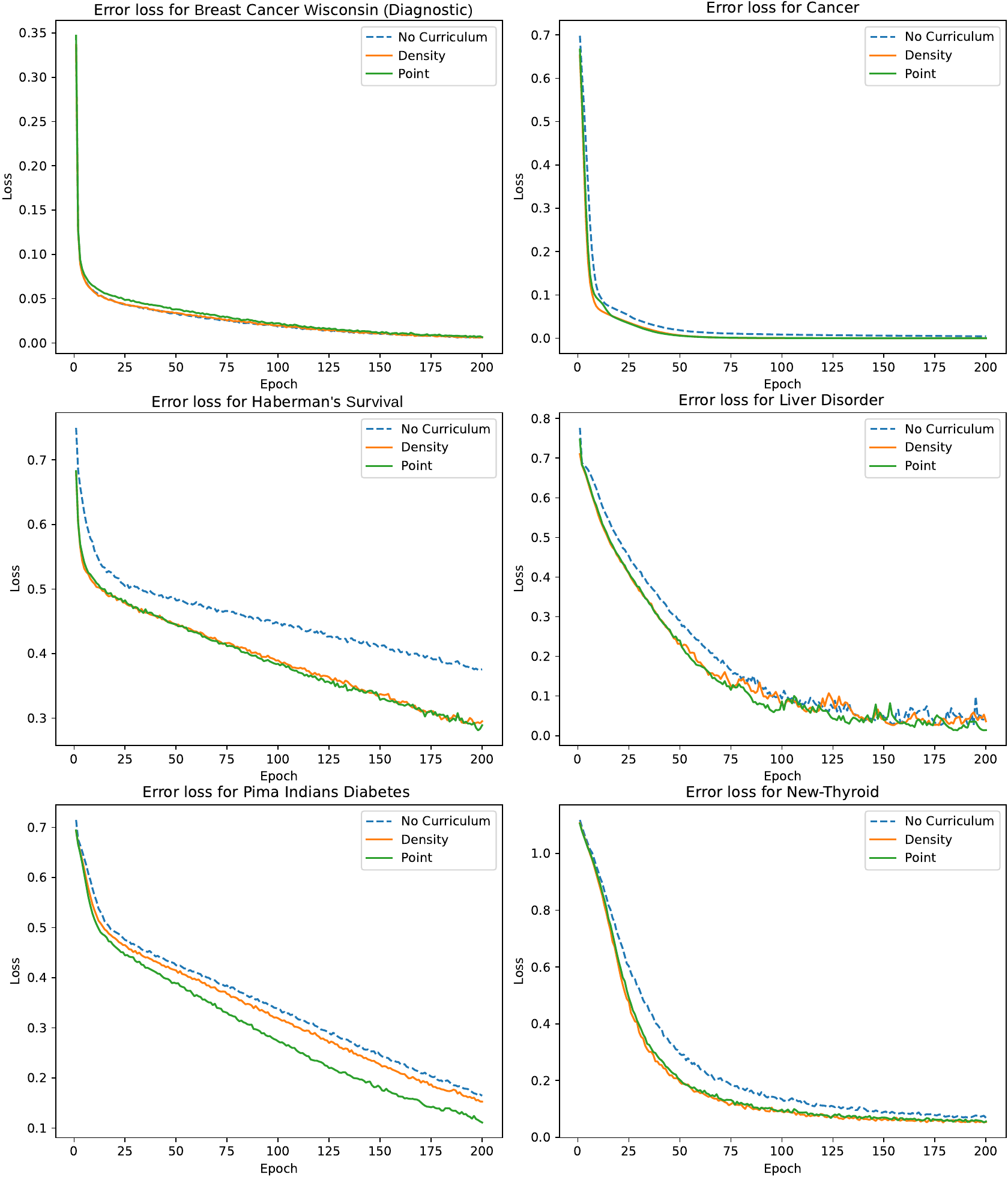}
    \caption{Error loss per epoch for neural network classifier.}
    \label{Fig:Loss_Epoch}
\end{figure*}

The results for neural networks, SVM and random forest classifiers are given in Table \ref{Table:NN_results}, Table \ref{Table:SVM_results} and Table \ref{Table:RF_results} respectively.

\section{Discussion}
\label{Section:Discussion}
Observations on the learning methods show that the classification performance across the datasets is always increased whenever DDCL is applied. This is true regardless of the learning method used as shown in Table \ref{Table:NN_results}, Table \ref{Table:SVM_results} and Table \ref{Table:RF_results}.
\begin{figure*}[ht]
    \centering
    \includegraphics[scale = 0.50]{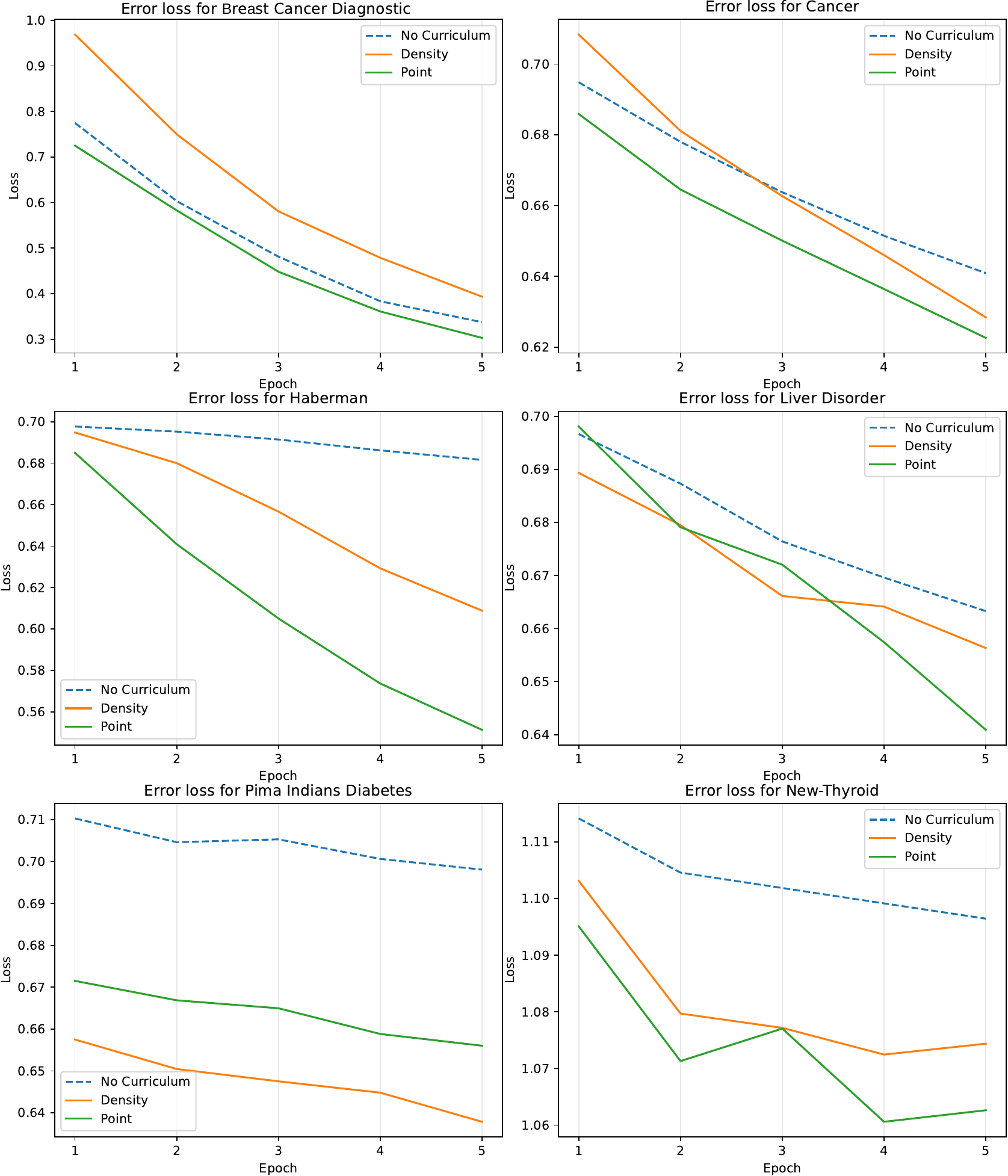}
    \caption{Error loss per epoch with batch gradient descent for neural network classifier.}
    \label{Fig:Loss_Batch}
\end{figure*}

Using the neural network classifier provides the highest best results for the Haberman's Survival, Liver Disorder and Pima Indians Diabetes datasets whereas the SVM method achieves the highest average results on the Breast Cancer (Diagnostic), Haberman's Survival and Pima Indians Diabetes data. The random forest based classifier yields the highest average result for the Cancer, Liver Disorder, New-Thyroid and Diabetes 130 datasets. Notably, using DDCL with SVM and random forest approaches results in significant increases in average accuracy for Haberman's Survival and Liver Disorder data respectively compared to other learning methods. Figure \ref{Fig:Precision_Recall} shows the precision-recall curves for the best result of each binary classification dataset when using the neural network method. For multi-class data, the confusion matrix is provided in place of the precision-recall curves by Figure \ref{fig:Confusion_Matrix}.

Table \ref{Table:DDCL_Compare} shows a comparison of results obtained using DDCL compared to approaches used by other state-of-the-art works. It demonstrates DDCL's ability to achieve superior results compared to the same type of classifier trained using a different algorithm. In particular it outperforms the balanced stratified method used in \cite{Zakaria2023_Pima_Diabetes_Haberman} on the challenging Haberman's Survival and Pima Indians Diabetes datasets. Furthermore, DDCL provides greater generalization capability and performance with neural networks on the Liver Disorder data compared to the standard approach used in \cite{Kumar2023_Liver_Disorder}.

The highest average result for Breast Cancer (Diagnostic) was achieved using SVM with density-based DDCL while the largest improvement (1.087\%) was using the neural network with point-based DDCL. For the Cancer dataset, using random forest with point-based DDCL provided the highest average accuracy and using the neural network with density-based DDCL gave the most improved accuracy (1.897\%). Using SVM with density-based DDCL resulted in the highest average accuracy when evaluated on the Haberman's Survival data. The largest improvement for this dataset (3.334\%) was observed when using density-based DDCL with the neural network. The highest average result for Liver Disorder was achieved using random forest with density-based DDCL whereas the SVM gave the largest average accuracy improvement at 4.808\%. For the Pima Indians Diabetes dataset, the highest average accuracy was obtained using SVM with density-based DDCL which was also the most improved accuracy (1.082\%).

In multi-class classification, using random forest with density-based DDCL resulted in the highest average accuracy on the New-Thyroid data and the largest improvement for this dataset (2.353\%) came from density-based DDCL applied to the neural network. For the Diabetes 130 data, density-based DDCL with random forest provided the highest average accuracy while the largest improvement in accuracy (2.550\%) was obtained through the neural network classifier utilizing density-based DDCL.

The experiment results demonstrate that using DDCL to order the training data for a given learning method leads to improvements in classification accuracy. This can be seen in the performance of all datasets tested with either the density or point-based DDCL approach having increased performance compared to the no curriculum approach. Figure \ref{Fig:Loss_Epoch} shows the error loss per epoch plots when using the neural network classifier for each dataset except for Diabetes 130 which is excluded since its average accuracy is not significant \cite{Shane2018_Readmission_Prediction}. The plot for Breast Cancer (Diagnostic) shows a minimal change in the DDCL error loss trends compared to the no curriculum method whereas the Cancer dataset plot presents a reduction in the error loss when using DDCL. For the remaining datasets, the use of DDCL shows a significant and quick reduction in the error loss.

In addition, using DDCL to order the training data results in faster convergence. We arrived at this conclusion by examining the error loss for the first five epochs of each dataset using batch gradient descent. Figure \ref{Fig:Loss_Batch} shows the error loss plots for each dataset. It can be seen that the overall losses for DDCL-Density and DDCL-Point are reduced faster compared to the no curriculum approach thus leading to faster convergence towards the minimum. This is consistent with the findings of the error loss against epoch plots in Figure \ref{Fig:Loss_Epoch} where the error loss towards the end is lower for DDCL-Density and DDCL-Point.

\section{Conclusion}
\label{Section:Conclusion}
This paper proposed a curriculum learning approach known as \textit{Data Distribution-based Curriculum Learning (DDCL)}. It used the data distribution of a dataset to build a curriculum based on the order of samples. Two types of scoring methods known as DDCL-Density and DDCL-Point were used to score samples thus determining their training order. DDCL-Density used the sample density to assign scores whereas DDCL-Point utilized the Euclidean distance for scoring.

Experiments were conducted on multiple medical datasets using a neural network, SVM and random forest classifier to evaluate DDCL. Results showed that even though performance varied across the datasets between the classifiers used, the application of DDCL resulted in accuracy increases ranging from 2\% to 10\% for all datasets compared to other state-of-the-art methods. Furthermore, analysis of the error losses for five training epochs using batch gradient descent reveals that convergence is faster when using DDCL with either of the scoring methods over the no curriculum method.

The current DDCL approach uses only two types of scoring methods and is limited to using a single scoring method at a time. Moreover, the proposed approach uses a fixed curriculum that is pre-determined before the start of training and does not take into account the current training progress.

Future work will explore the creation of additional scoring methods to determine their impact on the training performance and investigate the application of ensemble learning using each scoring method. By introducing these new scoring methods, we aim to assess their effectiveness in comparison to other curriculum approaches such as SPL and Reverse Curriculum Learning (RCL). In addition, self-paced learning concepts will be incorporated into DDCL in order to dynamically determine the curriculum based on feedback from the learner. This integration will allow the model to adjust its learning pace according to its current performance, potentially leading to a more robust and adaptive training process. Furthermore, the algorithms of the proposed DDCL's scoring methods suggests versatility which can be adapted beyond its initial application to structured datasets. Specifically, the proposed approach can also be adapted to image and text processing applications where the inherent complexity and variability of data types present unique challenges.

\section*{Nomenclature}

\begin{table}[ht]
    \begin{tabular}{|c|l|}
        \hline
        \textbf{Symbol} & \textbf{Description} \\
        \hline
        $S$ & Grouped data samples for all classes \\
        \hline
        $s$ & Data for a specific class \\
        \hline
        ${O_s}$ & Centroids for a specific class \\
        \hline
        $E_s$ & Euclidean distances from centroid (dimensionless) \\
        \hline
        $D_s$ & Data distribution for a class \\
        \hline
        $Q^s$ & Quantiles for the whole dataset \\
        \hline
        $q$ & A single quantile in a dataset \\
        \hline
        $R_s$ & Scored training data \\
        \hline
        $T$ & Curriculum prepared training data \\
        \hline
        $C_s$ & Densities (count) for all quantiles \\
        \hline
        $\hat{E_s}$ & Normalized Euclidean distances from centroid \\
        \hline
    \end{tabular}
\end{table}

\bibliographystyle{IEEEtran}
\bibliography{curriculum}

\pagebreak
\begin{IEEEbiographynophoto}{Shonal Chaudhry} received the M.S degree in computing science from The University of the South Pacific, Laucala Campus, Fiji in 2016.

From 2015 to 2021, he was a full stack developer. He is currently a PhD Researcher at The University of the South Pacific. His research interests are in artificial intelligence and its applications with a focus on machine learning, computer vision and curriculum-learning.
\end{IEEEbiographynophoto}

\begin{IEEEbiographynophoto}{Dr Anuraganand Sharma} (Senior Member, IEEE) received B.S. and M.S degrees in computer science from the University of the South Pacific, Fiji, and the Ph.D. degree in artificial intelligence from the University of Canberra, Australia, in 2014.

From 2003 to 2005, he was a software developer and since 2007, he joined the University of the South Pacific as an academic. His research interests are centered on deep learning with CNN and constraint optimization with meta-heuristic algorithms. His recent work includes SMOTified-GAN for class imbalanced problems and enhancement of SGD for gradient-based learning systems.
\end{IEEEbiographynophoto}

\EOD

\end{document}